\documentclass[10pt,twocolumn,letterpaper,table]{article}

\usepackage{iccv}              

\usepackage{lipsum}
\usepackage{array}
\usepackage{times}
\usepackage{epsfig}
\usepackage{graphicx}
\usepackage{float}
\usepackage{wrapfig}
\usepackage{amsmath,amssymb,amsthm}
\usepackage{algorithm,algorithmicx,algpseudocode}
\usepackage{bm,xspace}
\usepackage{comment}
\usepackage{multirow}
\usepackage{tabularx}
\usepackage{balance}
\usepackage{makecell}
\usepackage{url}
\usepackage{booktabs}
\usepackage{etoolbox,siunitx}
\usepackage{calc}
\usepackage{pifont,hologo}
\usepackage{color}
\usepackage{adjustbox}
\usepackage{amsmath}
\usepackage{enumitem}
\usepackage{caption}

\usepackage{bbding}  
\usepackage[normalem]{ulem}  
\usepackage{contour}
\usepackage[dvipsnames]{xcolor}
\usepackage{soul}
\usepackage{tocloft} 
\usepackage{etoc} 
\PassOptionsToPackage{dvipsnames}{xcolor}
\usepackage{natbib}
\PassOptionsToPackage{square,numbers,comma,sort}{natbib}

\newcommand\tightpara[1]{\noindent\textbf{#1}}

\definecolor{darkblue}{HTML}{35394B}
\definecolor{blue}{HTML}{004bb3}
\definecolor{red}{HTML}{cc1100}
\definecolor{orange}{HTML}{cc7700}
\definecolor{gray}{HTML}{efefef}
\definecolor{darkgreen}{HTML}{228B22}
\definecolor{darkgray}{HTML}{808080}
\definecolor{lightpurple}{HTML}{a56dba}

\definecolor{cite}{HTML}{3270b5}
\definecolor{link}{HTML}{b53532}
\definecolor{link}{HTML}{cc1100}
\definecolor{scratch}{HTML}{001219}
\definecolor{pretrain}{HTML}{0a9396}

\contourlength{0.8pt}

\renewcommand{\eqref}[1]{Eq.~\ref{#1}}

\newcolumntype{x}[1]{>{\centering\arraybackslash}p{#1}}
\newcolumntype{y}[1]{>{\raggedright\arraybackslash}p{#1}}
\newcolumntype{z}[1]{>{\raggedleft\arraybackslash}p{#1}}

\definecolor{verylightgray}{RGB}{245,245,245} 
\definecolor{pinkwrong}{HTML}{FE64A3}
\definecolor{greenright}{HTML}{46CB5F}
\newcolumntype{E}{>{\centering\arraybackslash}X}
\newcommand{\redxmark}{{\color{pinkwrong} \ding{55}}}
\newcommand{\greencheckmark}{{\color{greenright} \ding{51}}}

\DeclareMathSymbol{@}{\mathord}{letters}{"3B}

\makeatletter
\DeclareRobustCommand\onedot{\futurelet\@let@token\@onedot}
\def\@onedot{\ifx\@let@token.\else.\null\fi\xspace}

\newcommand*{\Rom}[1]{\expandafter\@slowromancap\romannumeral #1@}
\newcommand*{\rom}[1]{\expandafter\romannumeral #1}

\def\1{\bm{1}}

\DeclareMathAlphabet{\mathsfit}{\encodingdefault}{\sfdefault}{m}{sl}
\SetMathAlphabet{\mathsfit}{bold}{\encodingdefault}{\sfdefault}{bx}{n}

\let\originalleft\left
\let\originalright\right
\renewcommand{\left}{\mathopen{}\mathclose\bgroup\originalleft}
\renewcommand{\right}{\aftergroup\egroup\originalright}

\definecolor{iccvblue}{rgb}{0.21,0.49,0.74}
\usepackage[pagebackref,breaklinks,colorlinks,allcolors=iccvblue]{hyperref}

\def\paperID{10424} 
\def\confName{ICCV}
\def\confYear{2025}

\begin{document}
\title{Human-in-the-Loop Local Corrections of 3D Scene Layouts via Infilling}

\author{
Christopher Xie \quad
Armen Avetisyan \quad
Henry Howard-Jenkins \quad
Yawar Siddiqui \\
Julian Straub \quad
Richard Newcombe \quad
Vasileios Balntas \quad
Jakob Engel \\[6pt]
Meta Reality Labs \\[6pt]
\href{https://projectaria.com/scenescript}{https://projectaria.com/scenescript}
}

\twocolumn[{%
\renewcommand\twocolumn[1][]{#1}%
\maketitle
\vspace{-7.5mm}

\begin{center}
    \captionsetup{type=figure}
    \vspace{-2mm}
    \includegraphics[width=\linewidth]{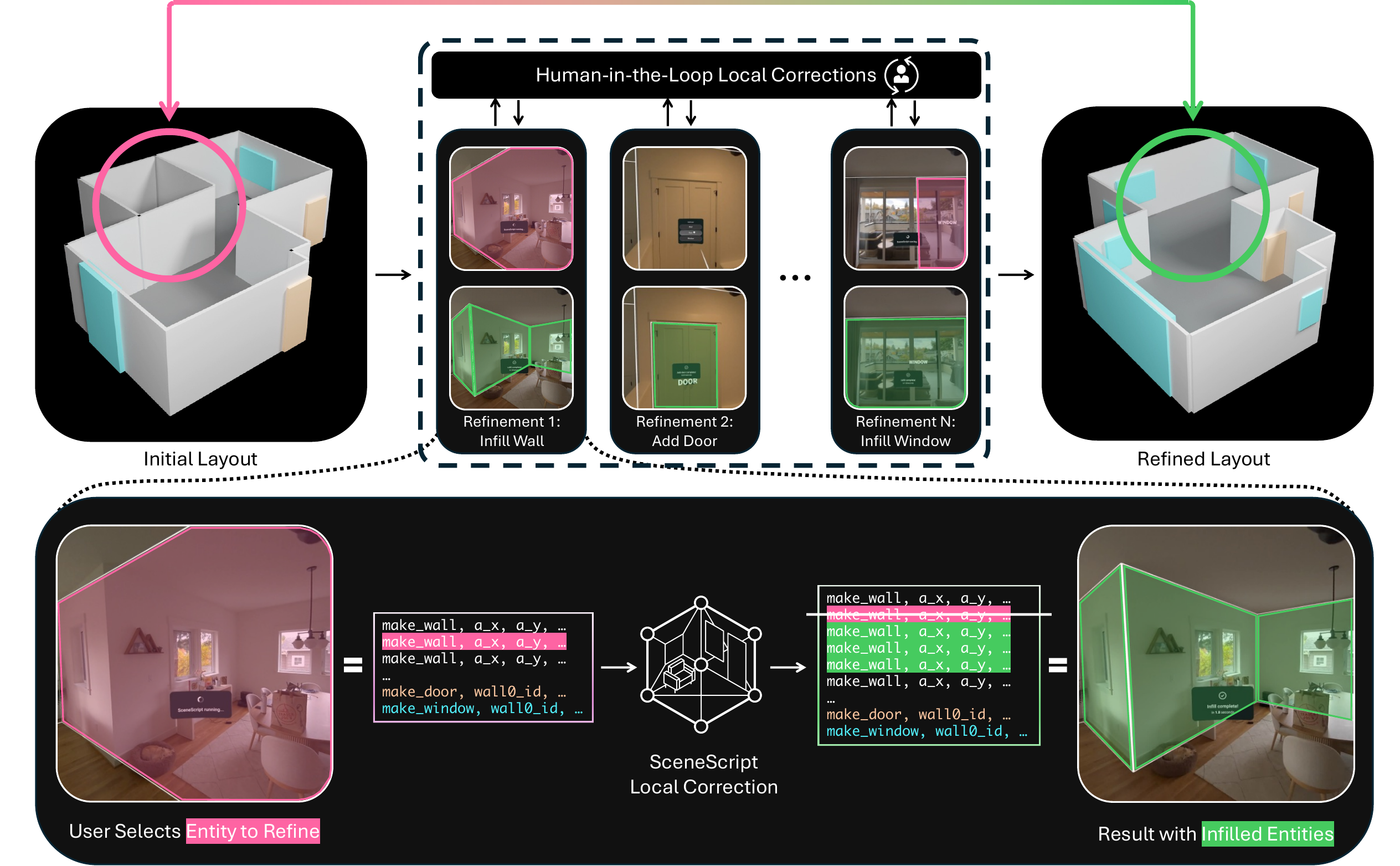}
    \captionof{figure}{Starting with an initial layout (top left) predicted by our model in global prediction mode, our system allows a user to identify \textcolor{pinkwrong}{local errors} within the layout. The user then prompts our model in local correction mode to refine these entities, resulting in \textcolor{greenright}{improved accuracy}. The user repeats this process (top center) until they achieve the desired layout (top right). (bottom) To perform the local correction, our method masks out the \textcolor{pinkwrong}{selected entities} and infills them with the \textcolor{greenright}{refined entities}. We highlight an incorrect wall structure in the initial layout (\textcolor{pinkwrong}{pink circle}) that is substantially improved (\textcolor{greenright}{green circle}) in the refined layout. Please see our \href{https://projectaria.com/scenescript}{project webpage} for videos of our system.}
    \label{fig:teaser}
\end{center}%
}]

\begin{abstract}
  We present a novel human-in-the-loop approach to estimate 3D scene layout that uses human feedback from an egocentric standpoint.
We study this approach through introduction of a novel local correction task, where users identify local errors and prompt a model to automatically correct them. 
Building on SceneScript~\cite{avetisyan2024scenescript}, a state-of-the-art framework for 3D scene layout estimation that leverages structured language, we propose a solution that structures this problem as ``infilling'', a task studied in natural language processing. 
We train a multi-task version of SceneScript that maintains performance on global predictions while significantly improving its local correction ability. 
We integrate this into a human-in-the-loop system, enabling a user to iteratively refine scene layout estimates via a low-friction ``one-click fix'' workflow.
Our system enables the final refined layout to diverge from the training distribution, allowing for more accurate modelling of complex layouts.

\end{abstract}

\section{Introduction}

Progress in 3D computer vision tasks such as reconstruction, object detection, and scene understanding has made notable strides, enabling more robust perception of complex environments. However, the prevailing paradigm in 3D computer vision relies on single-shot prediction models that produce a one-time, \textbf{global prediction} of visual and 3D data. 
These approaches inevitably encounter difficulties when applied to real-world scenarios, which are often out of distribution with respect to the training data due to the open-ended, ill-posed nature of these environments induced by partial observability and occlusions.
Instead, we adopt a human-in-the-loop paradigm grounded in a local egocentric perspective, making use of the user's agency within immersive 3D environments such as mixed reality. In this setting, we envision global predictions being iteratively refined locally by a model in response to prompts from a human, capitalizing on the human ability to detect model errors.

While interactive methods have seen success in 2D computer vision, such as with the Segment Anything Model (SAM)~\cite{kirillov2023segment}, applying a human-in-the-loop paradigm to 3D tasks remains largely unexplored. 
To explore human-in-the-loop paradigms in 3D, we focus on 3D scene layout estimation conducted from an egocentric context.
We build upon SceneScript~\cite{avetisyan2024scenescript}, a promising technique for scene layout estimation that leverages language modeling to perform a single-shot \textbf{global prediction} of 3D scene layout from an egocentric walkthrough.


To perform this iterative refinement for 3D scene layout, we introduce the novel task of \textbf{local correction} where a learned model must fix a local error in the scene layout estimate, which is identified by a user. We adapt SceneScript to simultaneously learn the task of local correction in addition to global prediction. By framing this task as "infilling"~\cite{donahue2020enabling,fried2022incoder,shen2023film,bavarian2022efficient}, which aims to generate text to fill in missing spans of a document, we leverage insights from natural language processing to develop a multi-task SceneScript model that can perform both \textbf{global prediction}s and \textbf{local correction}s. Interestingly, this multi-task approach maintains strong performance on global predictions compared to the original SceneScript model, while yielding significantly stronger results on local corrections.
We focus this investigation on architectural entities (walls/doors/windows) to prove the core concept of ``infilling''. As our method leverages SceneScript's~\cite{avetisyan2024scenescript} structured language, it can be easily extended to more general 3D scene layout including bounding boxes and/or volumetric primitives.


We integrate our model into a \textit{human-in-the-loop system} that allows a user to seamlessly refine scene layout estimates via a low-friction ``one-click fix'' workflow. 
Our system initializes the scene layout estimate using the model's \textbf{global prediction}.
Then, the user identifies (``one-click'') an error and triggers the same model's \textbf{local correction} capability (``fix'') to refine it. 
This process is iterated until the desired layout is achieved.
To immersively engage with 3D content, we utilize mixed reality technologies via a Meta Quest 3 headset with a Project Aria~\cite{aria_white_paper} rig rigidly attached on top.
Figure~\ref{fig:teaser} visualizes our interactive system (see supplemental video). 
Our iterative local corrections allow the \textit{final refined layout to diverge from the training distribution}, which is crucial in modeling complex real-world scenes.


To summarize, our core contributions are:

\begin{itemize}
    \item We introduce a human-in-the-loop paradigm for 3D scene layout estimation from an egocentric perspective, where human prompts guide iterative local model refinements in an interactive manner, addressing limitations of \textbf{global prediction} approaches in complex real-world settings.
    \item We define the novel task of \textbf{local corrections} and propose a solution based on infilling to yield a multi-task version of SceneScript.
    \item We show that simultaneously training SceneScript on global predictions \textit{and} local corrections results in maintained performance on the former task yet significantly stronger results on the latter task.
    \item We integrate our model into a human-in-the-loop system in a mixed reality environment employing a ``one-click fix'' workflow for seamless iterative refinements, enabling the final refined layout to diverge from the training distribution and effectively model complex real-world scenes. 
    Please see our \href{https://projectaria.com/scenescript}{project webpage} for videos of our system.
\end{itemize}

\section{Related Work}

\tightpara{Structured Language.}
Recently, several works have started exploring structured language representations to model geometry.
The seminal work of PolyGen~\cite{nash2020polygen} explored directly tokenizing mesh vertices and faces, and trained autoregressive Transformers~\cite{vaswani2017attention} to generate them.
MeshXL~\cite{chen2024meshxl} simplified the mesh representation with direct embeddings of the discretized nine scalar values that make up a mesh triangle. Additionally, they demonstrate that scaling the size of the model and data results in better stronger evaluation. 
MeshGPT~\cite{siddiqui2024meshgpt} uses six tokens per face. However, it first learns a codebook of quantized embedding with VQ-VAE~\cite{van2017neural}, then proceeds to learn an autoregressive Transformer to generate the learned token sequences.

Visual modeling software has also been considered in related research. 
CAD-as-Language~\cite{ganin2021computer} generates 2D computer-aided design (CAD) sketches with constraints in a similar fashion to PolyGen.
DeepCAD~\cite{wu2021deepcad} looks at generating 3D CAD sketches with a feedforward Transformer.
DeepSVG~\cite{carlier2020deepsvg} learns to generate vector graphics with a variational autoencoder.
BlenderAlchemy~\cite{huang2024blenderalchemy} leverages vision-language models (VLMs) to edit Blender~\cite{blender2023} programs from natural language descriptions.

Other works have designed their own structured language to represent geometry. Pix2Seq~\cite{chen2021pix2seq} directly tokenize 2D bounding box coordinates and generate them conditioned on an image.
SceneLanguage~\cite{zhang2024scene} represents scenes with program code to encode hierarchy, natural language to describe semantics of an entity, and embeddings to capture the visual identity of the entity.
SceneScript~\cite{avetisyan2024scenescript} proposes structured language commands to represent architectural layout, 3D bounding boxes, and volumetric primitives for objects. We build on top of SceneScript's custom language representation due to its ease of use and extensibility.

\tightpara{Text Infilling.} 
Infilling is a task in NLP that aims to fill in missing spans of text at any position of a document~\cite{donahue2020enabling}. 
It extends the cloze task~\cite{taylor1953cloze} whose objective is to fill in randomly deleted words from context. 
Masked language modeling such as BERT~\cite{devlin2018bert} and FiLM~\cite{shen2023film} employ this task for generative pre-training, using bidirectional context from Transformers as context. However, such methods can only generate tokens of a fixed length. 
To address this, ILM~\cite{donahue2020enabling} and CM3~\cite{aghajanyan2022cm3} proposed to generate the infilling answer at the end of the sentence, capitalizing on the autoregressive capabilities of Transformers.
Language models for code have also used infilling capabilities for editing code given bidirectional context~\cite{roziere2023code,fried2022incoder}.

``Fill In the Middle'' (FIM)~\cite{bavarian2022efficient} proposes a straightforward data augmentation in which an arbitrary span of the training text is moved to the end. By rearranging the document segments (prefix, middle, suffix) into (prefix, suffix, middle) and training with the standard next-token prediction loss, their language model effectively learns both strong left-to-right generative abilities and the ability to infill. Notably, when applying this augmentation with probability 0.5 to the training data, they observe that their language model learns to infill while keeping the original autogressive capability intact. They describe this as the ``FIM-for-free'' property. 
Building on this insight, we seek to develop a multi-task version of SceneScript that can perform multiple tasks while preserving its performance on the original task.







\tightpara{Interactive 2D Vision.}
A well-studied interactive computer vision problem is interactive image/video segmentation. Solutions in this space use user clicks to generate and refine object masks in a 2D image. If the result is inaccurate, additional clicks help correct errors and improve the segmentation. Our approach follows a similar collaborative philosophy.
Kontogianni et al.~\cite{kontogianni2020continuous} propose incorporating user corrections in an interactive system as a sparse training signal to update the underlying model on-the-fly. SimpleClick~\cite{liu2023simpleclick} enhances click-based interactive image segmentation by introducing a ViT backbone that effectively encodes user clicks into the model, achieving state-of-the-art segmentation performance.
FocalClick~\cite{chen2022focalclick} and FocusCut~\cite{lin2022focuscut} introduce a similar local refinement approach centered around the user's click to achieve high-quality segmentation.
SAM~\cite{kirillov2023segment} and SAM2~\cite{ravi2024sam} propose a promptable segmentation model ingesting user clicks to deliver final foreground object mask prediction. Due to the ViT architecture, a variable number of user clicks can be readily incoorporated leading to strong zero-shot performance on segmentation tasks via prompting.

\vspace{-2mm}
\section{Method}
\vspace{-1mm}


We use SceneScript's~\cite{avetisyan2024scenescript} structured language commands to represent scene layout. A hand-designed set of commands and parameters is used to describe geometric entities for scene layout, e.g. \texttt{make\_door} with parameters such as \texttt{height}, \texttt{width}. A scene layout can be described as a sequence of these commands, which can be converted into a sequence of tokens ready to be processed by a language model.
To investigate the core concept of ``infilling'', we focus only on architectural geometries in this work, i.e. walls, doors, and windows. Our method is agnostic to any structured language definition, thus it can directly extended to other commands (e.g. bounding boxes) via plug-and-play. 


We employ SceneScript's encoder-decoder architecture. The model consumes semi-dense point clouds (no images) and outputs a sequence of tokens that can be de-serialized into structured language commands. The encoder is a sparse 3D ResNet~\cite{tangandyang2023torchsparse} that produces features which are cross-attended to by a Transformer Decoder~\cite{vaswani2017attention}. Note that we do not introduce any changes to this architecture. 
Please refer to~\cite{avetisyan2024scenescript} for details about the structured language commands, custom tokenization procedure, and model details. 

\subsection{Tasks}

\begin{figure*}[t]
    \centering
    \captionsetup{type=figure}
    \includegraphics[width=\linewidth]{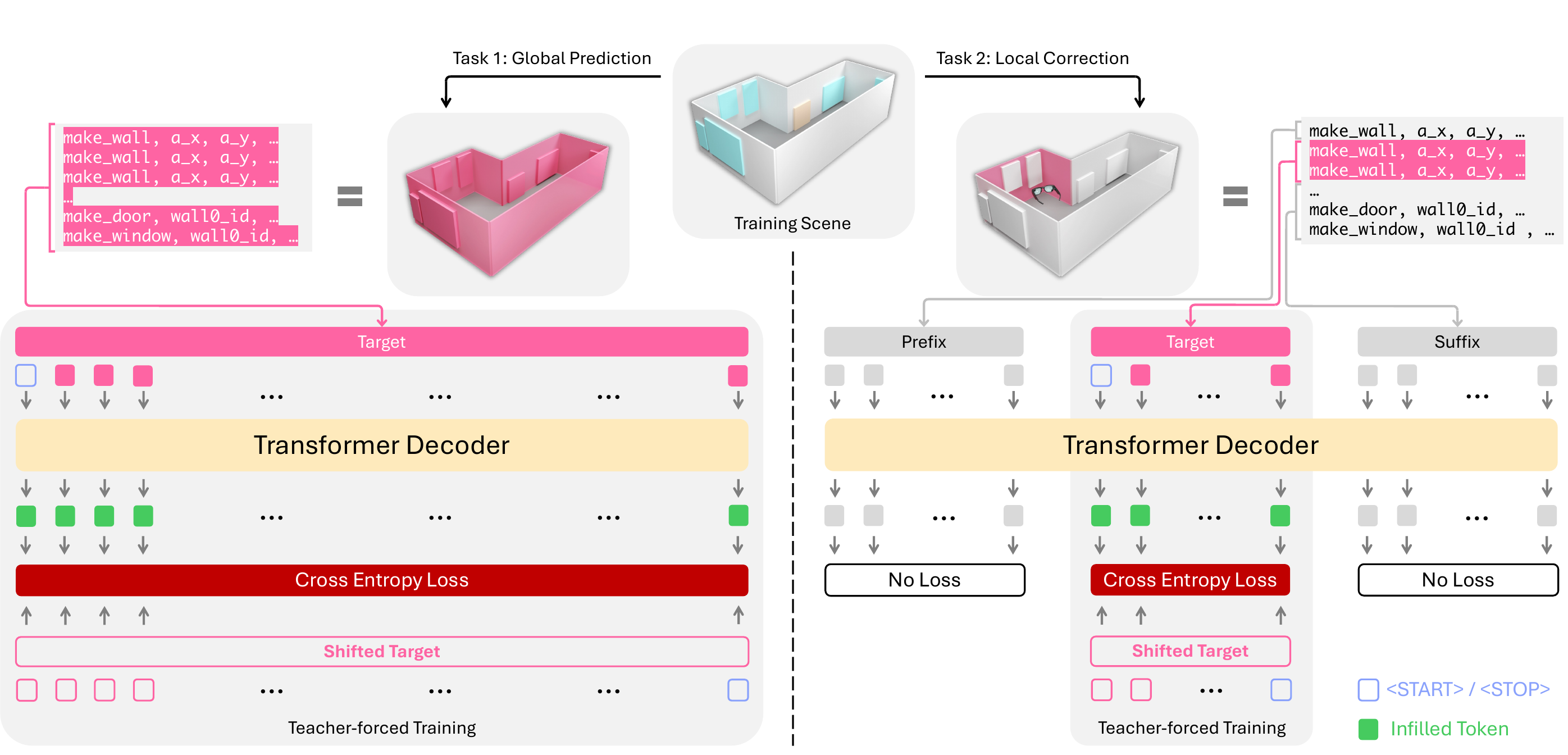}
    \caption{\textit{Global prediction} vs. \textit{local correction}. Given a training scene (top center), we choose with probability 0.5 whether the scene will be converted into a training sample for global prediction (left) or local correction (right). For global prediction (left), all layout entities are treated as \textcolor{pinkwrong}{targets}, and the decoder-only Transformer is trained via teacher-forcing to predict the entire scene. For local correction (right), a subset of layout entities are selected based on visibility from a given viewpoint (denoted by the Project Aria glasses). These entities are treated as \textcolor{pinkwrong}{targets} while the rest of the entities are sorted into prefix and suffix. Teacher-forcing is applied only to the selected entities. Note that semi-dense point clouds are cross-attended to in the Transformer, but are omitted in this illustration for brevity.}
    \label{fig:task_description_training}
    \vspace{-3mm}
\end{figure*}

Here we define our two tasks along with their training and inference procedures. We illustrate how SceneScript can accommodate both tasks in a single architecture. 

We assume access to a training dataset of indoor scenes, where each scene consists of an egocentric walkthrough of the scene with Project Aria glasses~\cite{aria_white_paper}, a semi-dense point cloud produced by running Project Aria's Machine Perception Services on the trajectory, and the corresponding ground truth layout in structured language format. Below, we describe how to construct training examples for both tasks from a single training scene.

\subsubsection{Global Prediction}

We refer to the scene layout estimation (via structured language) task introduced by SceneScript~\cite{avetisyan2024scenescript} as our global prediction task.  The details are as follows:

\tightpara{Inputs.} A semi-dense point cloud of the entire scene. 

\tightpara{Outputs.} A sequence of structured language tokens that represent the entire scene layout.

\tightpara{Training.} Here we recap the training setup described in SceneScript~\cite{avetisyan2024scenescript}: given a training scene, we first translate both the semi-dense point cloud and ground truth layout to the positive quadrant. Then, the structured language commands are sorted, continuous parameters are discretized, and the commands are tokenized into a 1D list of integers. As illustrated in Figure~\ref{fig:task_description_training} (left), the entire sequence is treated as the \texttt{target}, i.e. the network is trained to predict the next token for each token in the sequence. As with training standard autoregressive Transformers, teacher-forcing~\cite{williams1989learning} via cross entropy loss is applied to the whole sequence. 

\tightpara{Inference.} The semi-dense point cloud is first encoded and passed a conditioning signal to the decoder-only Transformer. Then, starting from a single \texttt{<START>} token, the decoder autoregressively predicts tokens until a \texttt{<STOP>} token is predicted. The integer token sequence is then de-serialized into structured language.

\subsubsection{Local Correction}
\label{subsubsec:local_correction}

We design this novel task with a low-friction ``one-click fix'' workflow in mind. In particular, we envision a human walking around in mixed reality with the current scene layout estimate. The user would then identify erroneous entities in the layout estimate with a single click, and trigger our model to automatically fix the selected entities.
In order to design this task as a \textit{local} correction, we assume that the selected entities are either 1) a set of connected walls that are currently visible by the user, 2) a single door, 3) or a single window.
With such a model, the user would be able to iteratively refine the scene layout estimate until the desired scene layout, which may be outside of the training distribution of indoor scenes, has been achieved.

Because we use SceneScript's structured language representation of scene layout, we can leverage the technical framework of ``infilling''~\cite{donahue2020enabling,fried2022incoder,shen2023film,bavarian2022efficient} as a solution to our task of local correction. Infilling seeks to predict missing spans of text at any position in a document. Similarly, our task seeks to predict missing spans of structured language. FIM~\cite{bavarian2022efficient} showed that jointly training \textit{a single Transformer} for both autoregressive generation and infilling results in retained performance for autoregressive generation and substantially increased performance for infilling.
We recognize the parallels between autoregressive generation and our task of global prediction; infilling and our task of local correction. Thus, we design local correction as an infilling task to leverage this insight.
The task is designed as follows:

\tightpara{Inputs.} 1) the 6DoF pose of the user's headset, 2) a semi-dense point cloud of the entire scene, and 3) the rest of the scene layout that has NOT been selected for correction. Importantly, the selected entities from the layout are excluded from the input, aligning with the infilling formulation.

\tightpara{Outputs.} A (short) sequence of structured language tokens that represent the corrected layout entities.

\tightpara{Training.} Given a training scene, we first sample a pose from the scene's egocentric trajectory and select entities that will serve as the \texttt{target}, i.e. the entities to infill. We consider only the layout entities that are visible from the pose (computed with Project Aria rig specifications~\cite{aria_white_paper}). An appropriate subset of entities that satisfies the assumptions mentioned above is randomly selected. 

After anchoring the scene at the user's pose instead of the positive quadrant, we perform the same operations as in the global prediction task (sort, discretize, tokenize). Only the selected entities serve as \texttt{target}, thus the network learns to \textit{infill} the selected entities while conditioned on the point cloud and the rest of the layout entities. These are split into a ``prefix'' or ``suffix'' based on ordering of the language commands, as shown in Figure~\ref{fig:task_description_training} (right). Teacher-forcing is applied only to the selected entities. Note that we add \texttt{<START>} and \texttt{<STOP>} tokens to the target subsequence.

\tightpara{Inference.} 
When a user has requested a local correction for selected entities, we construct the input as described in the training procedure. First, we center the current scene at the user's pose and compute the prefix/suffix subsequences. After encoding the semi-dense point cloud, we assemble the initial sequence from the prefix/suffix and a \texttt{<START>} token, then sequentially predict until it we hit \texttt{<STOP>}. The structured language can be de-serialized from the concatenated prefix, predicted target, and suffix subsequences.

\subsection{Design Choices}
\label{subsec:design_choices}

We shed light on our design choices here, which are ablated in Section~\ref{subsec:ablation_study}:

\tightpara{Joint Training.} 
The preceding section describes how a single SceneScript network can accommodate both tasks. Thus, we can train jointly on both tasks to yield a multi-task version of SceneScript. During training, we sample a global prediction training example with probability 0.5, otherwise we sample a local correction training example. Similarly to FIM~\cite{bavarian2022efficient}, we observe that SceneScript retains similar performance on global prediction while obtaining a substantial increase in performance for local corrections.

\tightpara{Language Ordering.} 
Structured language commands can be sorted in various ways. For example, sorting the commands randomly allows SceneScript (trained only for global prediction) to serve as a baseline for local corrections via partial layout completion~\cite{gupta2021layouttransformer}. We investigate ordering the commands randomly, lexicographically, or by angle.


Angle-based sorting is defined with respect to the user's pose.
As such, this sorting cannot be applied to global predictions where this is no notion of pose. In addition, we filter out entities that are not currently visible to the user in this ordering scheme. For a model trained jointly on both tasks with this sorting pattern, we use lexicographic sorting for global predictions, and angle sorting of visible entities for local corrections.
We show in Section~\ref{subsubsec:human-in-the-loop_system} that this particular ordering is useful for lowering the number of user interactions for iteratively refining a scene layout.

\tightpara{Subsequence Positional Embedding.} 
Instead of using the conventional learnable positional embedding~\cite{gehring2017convolutional} for absolute positions, we opt to give each prefix/target/suffix subsequence their own learnable positional embedding. This approach better leverages the segmented structure of the language commands.
Additionally, with this approach, SceneScript can sequentially generate an arbitrary number of tokens, unlike FiLM~\cite{shen2023film} which sequentially predicts a fixed number of tokens between the prefix/suffix.

\tightpara{Egocentric Anchoring.}
An important design choice is to exploit the user's context by centering the scene at the user's pose. Thus, selected entities are always in front of the user as opposed to anywhere in the scene, which makes the learning problem easier by lowering the entropy of the output token distributions.


\section{Experiments}
\label{sec:experiments}

\subsection{Implementation Details}

We use a slightly smaller version of SceneScript~\cite{avetisyan2024scenescript} with 4 Transformer decoder layers, 8 attention heads, and 512 model dimension. We make no changes to the sparse point cloud encoder. All models are trained for 500k iterations with cosine annealing~\cite{loshchilov2016sgdr}, a max learning rate of 5e-4, and an effective batch size of 64. We use the same augmentations as described in SceneScript for global predictions, and apply jittering in both translation (up to 20cm) and rotation (up to 20 degrees) of the user pose for local corrections.

When we anchor the scene at the user's pose, we maintain the gravity-aligned $z$ orientation of the world coordinate frame provided by Project Aria's Machine Perception Services~\cite{aria_white_paper} and only anchor the $xy$ orientation to the user's pose. For local correction (during training and inference), we can equivalently move the target to the end of the sequence (i.e. prefix, suffix, target) since each subsequence has their own positional embedding and due to the fact that Transformers are order-equivariant (note that FIM~\cite{bavarian2022efficient} does this as well). This allows us to capitalize on fast causal attention mechanisms~\cite{dao2023flashattention2}.

\subsection{Datasets, Metrics, and Baselines}
\label{subsec:datasets_metrics_baselines}

We train our models on Aria Synthetic Environments (ASE)~\cite{avetisyan2024scenescript}, a dataset consisting of 100k synthetic Manhattan-world indoor scenes. Each scene includes an egocentric scene walkthrough, a semi-dense point cloud, and ground truth layout as structured language. In the supplement, we train models on an internal proprietary version of ASE (iASE) that contains more complex synthetic layouts, thus better generalizes to real-world scenarios.

To measure out-of-distribution (OOD) generalization and real-world quantitative results, we use the Aria Everyday Objects (AEO)~\cite{straub24efm} dataset. This contains 25 challenging egocentric walkthroughs in real-world environments. The sequences are accompanied with manually-annotated scene layout. 
We note that while the style of layout annotation differs from that of ASE—for example, not all rooms are closed polygons—it still provides meaningful signal.

We quantitatively compare using the average F1 (AvgF1) metrics introduced in~\cite{avetisyan2024scenescript}, where higher is better. We use AvgF1 thresholds at every 5cm from 0m to 1m, which is slightly looser than the AvgF1 thresholds used in ~\cite{avetisyan2024scenescript}.

Due to the novelty of both task and system, there are \textbf{no existing baselines} for either. For local correction, we created a baseline using vanilla SceneScript (trained only on global predictions) via partial layout completion~\cite{gupta2021layouttransformer}: given all entities except the ones to correct, continue generating tokens until \texttt{<STOP>}. To our knowledge, no human-in-the-loop systems exist for low-friction editing of scene layouts.

\subsection{Ablation Study}
\label{subsec:ablation_study}

\begin{table}

\centering
\resizebox{\linewidth}{!}{%
\begin{tabular}{lccccccc}
\toprule
 & Eval & \multicolumn{3}{c}{Global Pred. AvgF1 $(\uparrow)$} &\multicolumn{3}{c}{Local Correction AvgF1 $(\uparrow)$} \\
 & Dataset & wall & door & window & wall & door & window \\
\cmidrule(lr){3-5} \cmidrule(lr){6-8}
Baseline~\cite{avetisyan2024scenescript} & ASE & \textbf{92.1} & \textbf{90.6} & \textbf{85.3} & 92.2 & 88.1 & 85.1  \\
\rowcolor{verylightgray} + joint training & ASE & 91.4 & 89.7 & 84.3 & \textbf{98.6} & \textbf{94.6} & \textbf{89.6}  \\
\midrule
Baseline~\cite{avetisyan2024scenescript} & AEO & \textbf{27.9} & \textbf{15.9} & 13.8 & 5.0 & 7.6 & 12.3  \\
\rowcolor{verylightgray} + joint training & AEO & 25.5 & 12.1 & \textbf{15.7} & \textbf{8.3} & \textbf{11.6} & \textbf{14.7}  \\
\cmidrule(lr){1-8}
\bottomrule
\end{tabular}%
}
\caption{Joint training ablation. We show AvgF1 scores for the global prediction (GP) and local correction (LC) tasks.}
\label{table:joint_training_ablation_asev1}
\vspace{-2mm}
\end{table}

\begin{table}

\centering
\resizebox{.65\linewidth}{!}{%
\begin{tabular}{lccc}
\toprule
& \multicolumn{3}{c}{Local Correction AvgF1 $(\uparrow)$} \\
Ordering & wall & door & window \\
\cmidrule(lr){2-4}
\rowcolor{verylightgray} lex & \textbf{99.1} & 94.4 & \textbf{90.5} \\
angle & 98.8 & 94.0 & 89.5 \\
\rowcolor{verylightgray} random & 98.6 & \textbf{94.6} & 89.6  \\
\bottomrule
\end{tabular}%
}

\caption{Language ordering ablation. }
\label{table:language_ordering_ablation_asev1}
\vspace{-2mm}
\end{table}

\begin{table}

\centering
\resizebox{.8\linewidth}{!}{%
\begin{tabular}{lclll}
\toprule
& & \multicolumn{3}{c}{Local Correction AvgF1 $(\uparrow)$} \\
Ordering & SPE & wall & door & window \\
\cmidrule(lr){3-5}

\multirow{2}{*}{lex} & \redxmark & 98.8 & 94.3 & 89.3 \\
 & \cellcolor{verylightgray} \greencheckmark & \cellcolor{verylightgray}99.1 {\footnotesize \textcolor{greenright}{(0.3)}} & \cellcolor{verylightgray}94.4 {\footnotesize \textcolor{greenright}{(0.1)}} & \cellcolor{verylightgray}90.5 {\footnotesize \textcolor{greenright}{(1.2)}} \\

\multirow{2}{*}{angle} & \redxmark & 98.5 & 93.4 & 88.9 \\
 & \cellcolor{verylightgray} \greencheckmark & \cellcolor{verylightgray}98.8 {\footnotesize \textcolor{greenright}{(0.3)}} & \cellcolor{verylightgray}94.0 {\footnotesize \textcolor{greenright}{(0.6)}} & \cellcolor{verylightgray}89.5 {\footnotesize \textcolor{greenright}{(0.6)}} \\
 
\multirow{2}{*}{random} & \redxmark & 98.7 & 94.4 &  89.4 \\
 & \cellcolor{verylightgray} \greencheckmark & \cellcolor{verylightgray}98.7 {\footnotesize (0.0)} & \cellcolor{verylightgray}94.9 {\footnotesize \textcolor{greenright}{(0.5)}} & \cellcolor{verylightgray}89.8 {\footnotesize \textcolor{greenright}{(0.4)}}\\
\bottomrule
\end{tabular}%
}
\caption{Subsequence positional embedding (SPE) ablation. }
\label{table:subsequence_positional_embedding_ablation_asev1}
\vspace{-2mm}
\end{table}

\begin{table}

\centering
\resizebox{.8\linewidth}{!}{%
\begin{tabular}{lclll}
\toprule
& & \multicolumn{3}{c}{Local Correction AvgF1 $(\uparrow)$} \\
Ordering & Ego & wall & door & window \\
\cmidrule(lr){3-5}
\multirow{2}{*}{lex} & \redxmark & 96.3 & 92.0 & 87.4 \\
 & \cellcolor{verylightgray} \greencheckmark & \cellcolor{verylightgray}98.8 {\footnotesize \textcolor{greenright}{(2.5)}} & \cellcolor{verylightgray}94.3 {\footnotesize \textcolor{greenright}{(2.3)}} & \cellcolor{verylightgray}89.3 {\footnotesize \textcolor{greenright}{(1.9)}} \\
\multirow{2}{*}{random} & \redxmark & 98.2 & 91.9 &  86.6 \\
& \cellcolor{verylightgray} \greencheckmark & \cellcolor{verylightgray}98.7 {\footnotesize \textcolor{greenright}{(0.5)}} & \cellcolor{verylightgray}94.4 {\footnotesize \textcolor{greenright}{(2.5)}} & \cellcolor{verylightgray}89.4 {\footnotesize \textcolor{greenright}{(2.8)}}\\
\bottomrule
\end{tabular}%
}
\caption{Egocentric anchoring ablation. }
\label{table:egocentric_anchoring_ablation_asev1}
\vspace{-5mm}
\end{table}

In this section, we study the design choices discussed in Section~\ref{subsec:design_choices}. We demonstrate their effects on both global prediction and local correction. Note that this section evaluates a single local correction using ground truth inputs, i.e. no iterative human-in-the-loop assistance. The goal is to assess the model's prediction capabilities in isolation.

\tightpara{Joint Training.} 
We examine whether joint training in our multi-task structured language setup results in a similar benefit as observed in~\cite{bavarian2022efficient}. These models are trained with random language ordering, and subsequence positional embeddings and egocentric anchoring for local corrections. 

For in-distribution evaluation on ASE (synthetic), Table~\ref{table:joint_training_ablation_asev1} shows that adding joint training results in a small dip in global prediction performance, around 1 point of AvgF1, but obtains a significant boost in local correction performance, between 4-7 points of AvgF1. This empirically demonstrates a similar result to FIM~\cite{bavarian2022efficient} in that training jointly on both tasks retains performance for the former task yet substantially increases performance on the latter task. We note that this small global prediction deterioration is consistent with other infilling works~\cite{roziere2023code, allal2023santacoder}.

Table~\ref{table:joint_training_ablation_asev1} shows similar trends for OOD generalization on AEO (real-world). While the global prediction performance is similar, the joint training clearly results in stronger local correction. However, the simplicity of scene layouts in ASE leads to poor and highly variable performance on AEO, primarily due to the substantial sim-to-real gap. Table~\ref{table:joint_training_ablation_asev2} in the supplement demonstrates that training on the more complex iASE with reduced sim-to-real gap results in improved performance while substantially amplifying the same trends.


\tightpara{Language Ordering.} 
We compare lexicographic, angle, and random language ordering in Table~\ref{table:language_ordering_ablation_asev1}. These models are trained with joint training, subsequence positional embeddings, and egocentric anchoring. The gap in local correction performance for these models is minimal again due to the simplicity of layouts in ASE. Additionally, the performance is nearly saturated. However, we can see that the lexicographic model slightly outperforms the angle model, which slightly outperforms the random model. For doors/windows, there is no obvious pattern, which correlates with the decision where only a single door/window is infilled at a time, thus ordering does not matter for the target. Table~\ref{table:language_ordering_ablation_asev2} in the supplement shows the same trends when training and evaluating with iASE but with a larger gap in performance due to the more complex layouts.


\tightpara{Subsequence Positional Embedding.} 
We compare using a per-subsequence positional embedding with a conventional positional embedding that embeds absolute position. These models are trained with joint training and egocentric anchoring. Table~\ref{table:subsequence_positional_embedding_ablation_asev1} demonstrates a consistent gain in performance on all local correction metrics when using subsequence positional embeddings, validating this design choice under various configurations. Table~\ref{table:subsequence_positional_embedding_ablation_asev2} shows this for iASE.

\tightpara{Egocentric Anchoring.}
Lastly, we study the effectiveness of anchoring the scene at the user's pose (for local corrections only) as opposed to translating the scene to the positive quadrant. These models are trained with joint training and conventional positional embeddings. Table~\ref{table:egocentric_anchoring_ablation_asev1} (and Table~\ref{table:egocentric_anchoring_ablation_asev2} in the supplement) confirms that egocentric anchoring results in improved accuracy across different parameter settings. Note that we do not ablate with angle sorting as this method requires egocentric anchoring.
\\
\\
We note that performance is essentially saturated for the local correction metrics. This is due to the simplicity of ASE (please see visual examples of ASE in~\cite{avetisyan2024scenescript}). However, in the supplement, we provide a more in-depth analysis of results, and study the same questions using iASE. We see the same trends with larger discrepancies in performance due to the more challenging scene layouts.

\subsection{Iterative Local Refinements}

In this section we describe how to integrate our model into a iterative ``one-click fix'' system. Given our multi-task version of SceneScript, we can utilize its global prediction mode to provide an initial scene layout estimate. We can then iteratively refine that estimate by performing actions that leverage the model's local correction capabilities. These actions are defined below:
\begin{enumerate}
    \item \texttt{Infill}: This is the local correction setup as in Section~\ref{subsubsec:local_correction}. Erroneous entities are identified (this is the ``one-click''), then locally corrected by the model (``fix'').
    \item \texttt{Add}: We leverage the above action for this: by fabricating a layout entity of a desired class directly in front of the user and applying the \texttt{Infill} action to it, we can effectively add a new entity of that class to the scene.
    \item \texttt{Delete}: We simply remove the layout entity from the structured language.
\end{enumerate}
These actions allow a user to iteratively refine the scene layout via a low-friction ``one-click fix'' workflow until the desired accuracy has been achieved. Note that all of our models have been trained purely in simulation, while all data used in this section is real-world. We demonstrate the efficacy of our system with two applications.

\subsubsection{Automatic Heuristic Local Corrections}
\label{subsubsec:automatic_heuristic_local_corrections}

\begin{figure}[t]
    \centering
    \captionsetup{type=figure}
    \includegraphics[width=\linewidth]{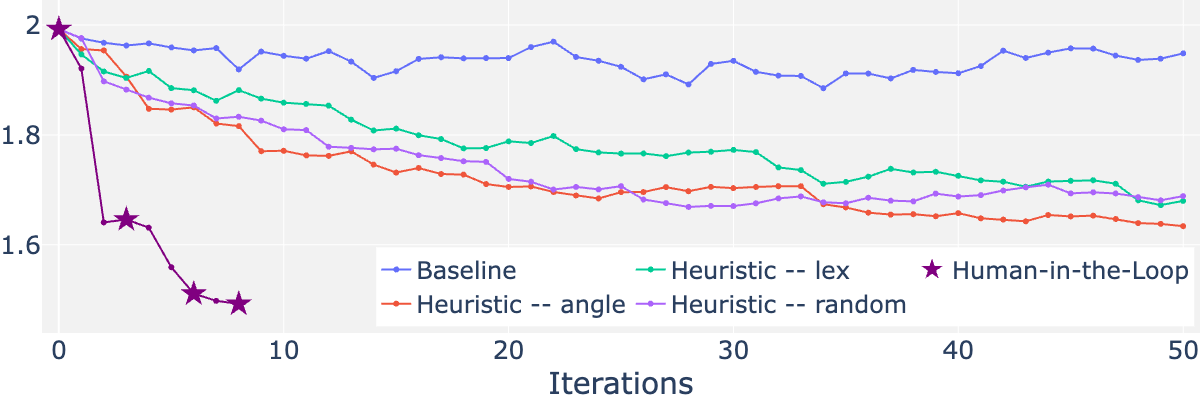}
    \caption{Average PlaneDistance $(\downarrow)$ results for heuristic and human-in-the-loop local corrections on AEO~\cite{straub24efm}.}
    \label{fig:aeo_heuristic_local_corrections_results_asev1}
    \vspace{-4mm}
\end{figure}

We first demonstrate this system with a heuristic to select erroneous entities instead of a user. Given a current scene layout estimate and a randomly selected frame from an Aria trajectory from AEO, we run Mask2Former~\cite{cheng2021mask2former} to obtain wall/door/window instance segmentation masks. Comparing these with the current scene layout estimate, we design a heuristic (see Algorithm~\ref{alg:action_selection_heuristic} in the supplement) to output a list of possible actions and their corresponding selected entities. We then select an action (and its corresponding entities) at random, execute it, and repeat this process.

Starting from the same initial layout estimate, we run the loop for 5 trials with 50 iterations each, and show Average PlaneDistance (lower is better) results averaged over the 5 trials, all 25 AEO sequences, and all classes (wall/door/window). 
Because the real-world problem is harder than simulation, we opt for PlaneDistance, i.e. the distance function $d_E(\cdot)$ defined in~\cite{avetisyan2024scenescript}, as opposed to AvgF1.
At every iteration, we compute PlaneDistance of the current layout compared to ground truth.
In Figure~\ref{fig:aeo_heuristic_local_corrections_results_asev1}, we show results of a SceneScript baseline and our model trained with different language orderings integrated the iterative system. All of our multi-task models show a trend of improvement as opposed to the baseline, which stays relatively flat.

While this heuristic iterative system shows improvement, Figure~\ref{fig:aeo_heuristic_local_corrections_results_asev1} also shows that using a human in the loop instead results in faster and better convergence of the scene layout. Note that the human-in-the-loop result uses the same exact model as the heuristic variant ``Heuristic -- lex''. This demonstrates that the heuristic is not as effective as a human in selecting actions. Future work includes minimizing this gap to further lower human intervention.

\subsubsection{Human-in-the-Loop System}
\label{subsubsec:human-in-the-loop_system}

\begin{figure*}[t]
    \centering
    \captionsetup{type=figure}
    \includegraphics[width=\linewidth]{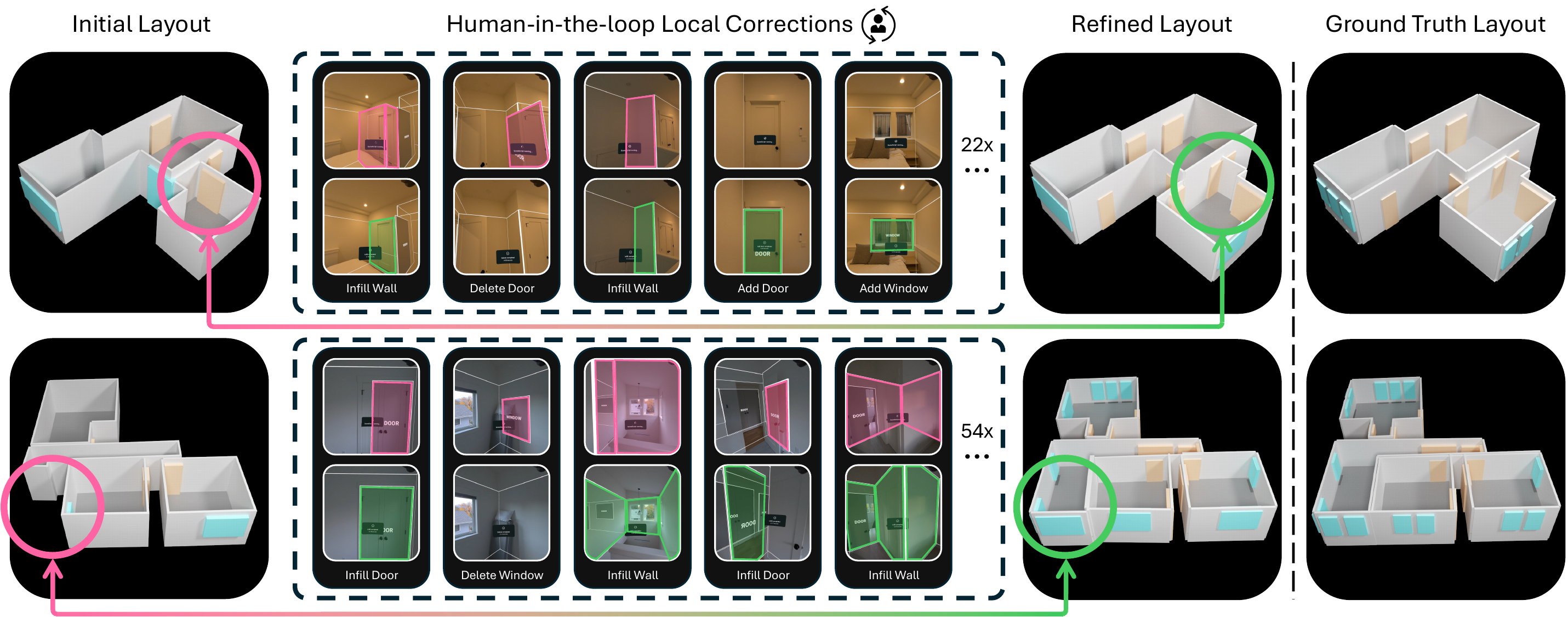}
    \caption{Snapshots from our human-in-the-loop system. We show the initial layout, a few local correction steps, and the final refined layout for two real-world scenes. For reference, we also visualize manually annotated ground truth. The center loop shows 5 local corrections for each scene with actions such as \texttt{Infill}/\texttt{Add}/\texttt{Delete} for walls/doors/windows. User-identified \textcolor{pinkwrong}{incorrect entities} are refined to \textcolor{greenright}{improved entities}. We highlight large errors in the initial layout (\textcolor{pinkwrong}{pink circles}) that are effectively remedied (\textcolor{greenright}{green circle}) in the refined layout.}
    \label{fig:human-in-the-loop_system}
    \vspace{-4mm}
\end{figure*}

Our second application includes a human in the loop to select erroneous entities (via a single click). This system uses our multi-task version of SceneScript trained with angle-based language ordering on iASE (for stronger real-world generalization) with one extra capability: when walls are sorted based on angle w.r.t. the user's pose, the last prefix and first suffix walls are connected to the selected walls (assuming all walls are connected to at least one other wall). Thus, we can additionally infill the last prefix and first suffix corners, which leads to fewer user interactions. For example, modifying the room layout from square to rectangle only requires a single local correction.


After scanning the scene with a Meta Quest 3 with an Aria rigidly attached on top to obtain an initial semi-dense point cloud, the user runs our model in global prediction model to obtain an initial layout estimate. The user can then select erroneous layout entities by clicking on them with the Quest 3 controllers and request an \texttt{Infill/Add/Delete} action. User pose is obtained via Project Aria Machine Perception Services~\cite{aria_white_paper}.

Figure~\ref{fig:human-in-the-loop_system} illustrates snapshots of our human-in-the-loop live system in two real-world scenes. We additionally manually label ground truth for these two scenes for qualitative purposes only. Figure~\ref{fig:teaser} also shows an example of a third scene. We highlight the following observations:
\begin{itemize}
    \item Using the \texttt{Infill} action, we can replace any number of walls with an arbitrary number of walls. In Figure~\ref{fig:human-in-the-loop_system} row 1 column 1, we infill two walls with one. In the first local correction of Figure~\ref{fig:teaser}, we infill one wall with three.
    \item Our system demonstrates successful \texttt{Add}-ing of doors and windows, e.g. see row 1, columns 4 and 5. Additionally, we can delete them as well, e.g. see column 2 of both rows. Note that we do not add walls due to the assumption that the last prefix wall and first suffix wall connect to the selected wall, which is violated when we fabricate a wall directly in front of the user.
    \item Alongside significant layout corrections, e.g. row 2, columns 3 and 5, our system can also perform minor alignment adjustments, as shown in the snapshots in row 1, column 3, and row 2, column 1. The system is further capable of relocating a door from an incorrect wall to the correct adjacent wall, as seen in row 2, column 4.
    \item The number of human interactions is 22 and 54 for the two scenes in Figure~\ref{fig:human-in-the-loop_system} (and 17 for the scene in Figure~\ref{fig:teaser}). The time per local correction is less than one second, delivering a responsive interaction and seamless user experience. Future work involves lowering the number of interactions required to achieve a desired accuracy.
    \item Our live system accumulates semi-dense points over time, enabling users to re-scan poorly captured areas and provide additional signal for local corrections. In the supplementary material, we show that even with increased point density, global prediction still fails to achieve the desired layout as it is typically OOD w.r.t. the training distribution. This highlights the advantages of our local iterative approach over single-shot global prediction methods. The idea of locality-for-generality has also proven effective in prior work on dense 3D reconstruction~\cite{jiang2020local,chabra2020deepls}.
\end{itemize}
As there is no existing baseline system, we conducted a user study comparing the integration of the vanilla SceneScript baseline vs. our model in our human-in-the-loop system. Please refer to the supplement for details and results, and the \href{https://projectaria.com/scenescript}{project webpage} for videos of our system.


\vspace{-2mm}
\section{Discussion and Limitations}
\vspace{-2mm}

Our approach demonstrates strong performance in predicting and correcting room structures. Because we build on top of SceneScript's structured language, we can easily adapt to new structures, including 3D bounding boxes and volumetric primitives, by updating the language schema. By design, we do not use selected entities as input during local correction, facilitating simple plug-and-play extensions.

Existing semi-automatic layout annotation tools, such as PanoAnnotator~\cite{yang2018panoannotator}, rely on continuous-value edits that demand significant user effort. Our method instead introduces a lower-friction “one-click fix” paradigm. Enabling users to correct geometric errors with a single click is non-trivial, posing an interesting problem our work seeks to address.



\vspace{-2mm}
\section{Conclusion}
\vspace{-2mm}

In this work, we introduced a human-in-the-loop method for 3D scene layout estimation that harnesses the user’s ability to identify model errors within an egocentric context.
We introduced a new task, local corrections, and proposed a solution based on infilling.
By adapting SceneScript to support both global and local predictions, we obtained a multi-task version that performs well for both tasks. 
We built a human-in-the-loop system that utilizes a low-friction ``one-click fix'' workflow, enabling the refined layout to diverge from the training distribution for high-fidelity 3D scene modelling.
Our approach suggests new pathways for interactive systems for 3D vision and mixed reality applications.

{
\clearpage
\small
\bibliographystyle{ieeenat_fullname}
\bibliography{main}
}

\clearpage
\appendix
\section*{Appendix}



\section{Additional Quantitative Results}

\subsection{Results on Internal Version of ASE}

\begin{table}

\centering
\resizebox{\linewidth}{!}{%
\begin{tabular}{lccccccc}
\toprule
 & Eval & \multicolumn{3}{c}{Global Pred. AvgF1 $(\uparrow)$} &\multicolumn{3}{c}{Local Correction AvgF1 $(\uparrow)$} \\
 & Dataset &wall & door & window & wall & door & window \\
\cmidrule(lr){3-5} \cmidrule(lr){6-8}
Baseline~\cite{avetisyan2024scenescript} & iASE & \textbf{79.4} & \textbf{87.2} & \textbf{77.7} & 87.2 & 88.0 & 83.1  \\
\rowcolor{verylightgray} + joint training & iASE & 77.5 & 87.0 & 77.5 & \textbf{92.5} & \textbf{92.9} & \textbf{88.0} \\
\midrule
Baseline & AEO & \textbf{34.9} & \textbf{27.8} &  \textbf{22.9} & 29.0 & 9.4 & 9.5  \\
\rowcolor{verylightgray} + joint training & AEO & 33.0 & 25.1 & 18.6 & \textbf{56.0} & \textbf{23.4} & \textbf{25.1}  \\
\bottomrule
\end{tabular}%
}
\caption{Joint training ablation. These models are trained on our internal proprietary version of ASE (iASE). We show AvgF1 scores for the global prediction (GP) and local correction (LC) tasks. Note that the gaps in OOD performance on AEO are substantially amplified compared to Table~\ref{table:joint_training_ablation_asev1} of the main paper.}
\label{table:joint_training_ablation_asev2}

\end{table}

\begin{table}

\centering
\resizebox{.65\linewidth}{!}{%
\begin{tabular}{lccc}
\toprule
& \multicolumn{3}{c}{Local Correction AvgF1 $(\uparrow)$} \\
Ordering & wall & door & window \\
\cmidrule(lr){2-4}
\rowcolor{verylightgray} lex & \textbf{93.4} & \textbf{93.5} & \textbf{89.7} \\
angle & 92.9 & 92.9 & 88.3  \\
\rowcolor{verylightgray} random & 92.5 & 92.9 & 88.0  \\
\bottomrule
\end{tabular}%
}

\caption{Language ordering ablation for models trained on iASE. }
\label{table:language_ordering_ablation_asev2}

\end{table}

\begin{table}

\centering
\resizebox{.8\linewidth}{!}{%
\begin{tabular}{lclll}
\toprule
& & \multicolumn{3}{c}{Local Correction AvgF1 $(\uparrow)$} \\
Ordering & SPE & wall & door & window \\
\cmidrule(lr){3-5}

\multirow{2}{*}{lex} & \redxmark & 92.7 & 92.9 & 88.3 \\
 & \cellcolor{verylightgray} \greencheckmark & \cellcolor{verylightgray}93.4 {\footnotesize \textcolor{greenright}{(0.7)}} & \cellcolor{verylightgray}93.5 {\footnotesize \textcolor{greenright}{(0.6)}} & \cellcolor{verylightgray}89.7 {\footnotesize \textcolor{greenright}{(1.4)}} \\

\multirow{2}{*}{angle} & \redxmark & 92.6 & 91.8 & 87.3 \\
 & \cellcolor{verylightgray} \greencheckmark & \cellcolor{verylightgray}92.9 {\footnotesize \textcolor{greenright}{(0.3)}} & \cellcolor{verylightgray}92.9 {\footnotesize \textcolor{greenright}{(1.1)}} & \cellcolor{verylightgray}88.3 {\footnotesize \textcolor{greenright}{(1.0)}} \\

\multirow{2}{*}{random} & \redxmark & 91.5 & 92.7 & 88.1 \\
 & \cellcolor{verylightgray} \greencheckmark & \cellcolor{verylightgray}92.5 {\footnotesize \textcolor{greenright}{(1.0)}} & \cellcolor{verylightgray}92.9 {\footnotesize \textcolor{greenright}{(0.2)}} & \cellcolor{verylightgray}88.0 {\footnotesize \textcolor{pinkwrong}{(0.1)}}\\

\bottomrule
\end{tabular}%
}
\caption{Subsequence positional embedding (SPE) ablation for models trained on iASE. }
\label{table:subsequence_positional_embedding_ablation_asev2}

\end{table}

\begin{table}

\centering
\resizebox{.8\linewidth}{!}{%
\begin{tabular}{lclll}
\toprule
& & \multicolumn{3}{c}{Local Correction AvgF1 $(\uparrow)$} \\
Ordering & Ego & wall & door & window \\
\cmidrule(lr){3-5}
\multirow{2}{*}{lex} & \redxmark & 92.9 & 92.7 & 86.6 \\
 & \cellcolor{verylightgray} \greencheckmark & \cellcolor{verylightgray}92.7 {\footnotesize \textcolor{pinkwrong}{(0.2)}} & \cellcolor{verylightgray}92.9 {\footnotesize \textcolor{greenright}{(0.2)}} & \cellcolor{verylightgray}88.3 {\footnotesize \textcolor{greenright}{(1.7)}} \\

\multirow{2}{*}{random} & \redxmark & 91.2 & 92.0 & 86.3 \\
& \cellcolor{verylightgray} \greencheckmark & \cellcolor{verylightgray}91.5 {\footnotesize \textcolor{greenright}{(0.3)}} & \cellcolor{verylightgray}92.7 {\footnotesize \textcolor{greenright}{(0.7)}} & \cellcolor{verylightgray}88.1 {\footnotesize \textcolor{greenright}{(1.8)}}\\
\bottomrule
\end{tabular}%
}
\caption{Egocentric anchoring ablation for models trained on iASE. }
\label{table:egocentric_anchoring_ablation_asev2}

\end{table}

In this section, we repeat the experiments in Section~\ref{subsec:ablation_study} with our internal version of ASE~\cite{avetisyan2024scenescript} (iASE) that contains more complex scene layouts.

\tightpara{Joint Training.} 
As observed in~\cite{bavarian2022efficient,roziere2023code,allal2023santacoder} and Table~\ref{table:joint_training_ablation_asev1}, we again see in Table~\ref{table:joint_training_ablation_asev2} a slight drop in performance on global prediction AvgF1 but a large gain on local correction AvgF1 when evaluating on in-distribution synthetic validation data (iASE). These models are trained with random language ordering, and subsequence positional embeddings and egocentric anchoring for local corrections.

However, on OOD validation data (AEO~\cite{straub24efm}), the gains on local correction AvgF1 are significantly amplified compared to the results in Table~\ref{table:joint_training_ablation_asev1} (trained on ASE, not iASE). The wall AvgF1 is nearly doubled in this setting, and the door/window AvgF1 are roughly 2.5x better than the baseline. Thus, training our models on the more complex iASE, which has less sim-to-real gap, provides strong evidence of the benefit of jointly training for both global prediction and local correction.

\tightpara{Language Ordering.} 
We compare lexicographic, angle, and random language ordering in Table~\ref{table:language_ordering_ablation_asev2}. These models are trained with joint training, subsequence positional embeddings, and egocentric anchoring. As in Table~\ref{table:language_ordering_ablation_asev1}, we see that the lexicographic model slightly outperforms the angle model, which slightly outperforms the random model. Gaps performance are slightly larger on this more complex dataset.

\tightpara{Subsequence Positional Embedding.} 
We compare using a per-subsequence positional embedding against a conventional positional embedding that embeds absolute position. These models are trained with joint training and egocentric anchoring. As seen in Table~\ref{table:subsequence_positional_embedding_ablation_asev2}, using the per-subsequence positional embedding shows a consistent gain in performance, corroborating the results in Table~\ref{table:subsequence_positional_embedding_ablation_asev1}.

\tightpara{Egocentric Anchoring.}
Table~\ref{table:egocentric_anchoring_ablation_asev2} studies the effectiveness of anchoring the scene at the user's pose (for local corrections only) as opposed to translating the scene to the positive quadrant. These models are trained with joint training and conventional positional embeddings. Results show the efficacy of this design choice across almost all metrics in multiple configurations of the models, similar to Table~\ref{table:egocentric_anchoring_ablation_asev1}.

\subsection{Multi-Wall Evaluation}

We note that in our evaluation of local corrections for walls, many of the examples only contain a single wall to infill. This problem is easy, as the corners for the missing wall are present in the prefix/suffix sequences, so the network only needs to learn to copy those corners from the correct places. When the number of walls $n_w > 1$, the network must predict the location of at least one corner. 

In Tables~\ref{table:language_ordering_ablation_asev1_multiwall} and~\ref{table:language_ordering_ablation_asev2_multiwall}, we show an analysis of this breakdown by $n_w$ for our models trained with different language orderings on ASE~\cite{avetisyan2024scenescript} and our internal version of ASE, respectively. We separate the evaluation from 1 to $\geq 4$ walls.
In both tables, performance on $n_w=1$ is nearly perfect as the problem is easy. However, the performance of all models degrades as $n_w$ increases, because the problem becomes more difficult. Similarly, the performance gaps for the methods generally increase as the number of walls increases.

\begin{table}

\centering
\begin{tabular}{lcccc}
\toprule
& \multicolumn{4}{c}{Local Correction AvgF1 $(\uparrow)$} \\
Ordering & $n_w=1$ & $n_w=2$ & $n_w=3$ & $n_w \geq 4$ \\
\cmidrule(lr){2-5}
\rowcolor{verylightgray} lex & \textbf{99.9} & \textbf{97.6} & \textbf{96.0} & \textbf{95.9} \\
angle & 99.8 & 96.9 & 95.1 & 95.4 \\
\rowcolor{verylightgray} random & 99.8 & 96.0 & 95.2 & 94.8 \\
\bottomrule
\end{tabular}%

\caption{Language ordering ablation breakdown by number of walls ($n_w$) on ASE~\cite{avetisyan2024scenescript}. All AvgF1 numbers are shown for walls only. }
\label{table:language_ordering_ablation_asev1_multiwall}

\end{table}

\begin{table}

\centering
\begin{tabular}{lcccc}
\toprule
& \multicolumn{4}{c}{Local Correction AvgF1 $(\uparrow)$} \\
Ordering & $n_w=1$ & $n_w=2$ & $n_w=3$ & $n_w \geq 4$ \\
\cmidrule(lr){2-5}
\rowcolor{verylightgray} lex & 99.4 & \textbf{93.3} & \textbf{88.8} & \textbf{84.3}  \\
angle & \textbf{99.5} & 92.1 & 87.9 & 81.8 \\
\rowcolor{verylightgray} random &  \textbf{99.5} & 91.7 & 86.6 & 80.1  \\
\bottomrule
\end{tabular}%

\caption{Language ordering ablation breakdown by number of walls ($n_w$) on our internal version of ASE. All AvgF1 numbers are shown for walls only.}
\label{table:language_ordering_ablation_asev2_multiwall}

\end{table}

\section{User Study}

Due to the novelty of the local correction task and our proposed human-in-the-loop system, there are no existing baseline systems to compare to. Thus, to obtain insight into our system, we conducted a user study comparing different models integrated into our low-friction ``one-click fix'' system. In particular, we compare the vanilla SceneScript baseline vs. our multi-task SceneScript (lexicographically-ordered), both trained on iASE. Note that these models have not been trained with the extra capability of infilling prefix/suffix corners described in Section~\ref{subsubsec:human-in-the-loop_system}.

We used pre-recorded Aria trajectories in real-world home and office environments. We used our system in an offline fashion by manually selecting a frame and the entities to fix. At each local correction step, we ran both the baseline model and our multi-task model, and asked users to select the local correction that they preferred. Averaging over 10 users and 18 local correction examples, 57.2\% preferred our model, 18.9\% preferred the baseline, and 23.9\% preferred neither. This highlights that the joint training is crucial in learning a model that can be integrated with our human-in-the-loop system.

We note that a common failure mode of the vanilla SceneScript baseline is to re-predict scene layout elements (walls/doors/windows) that already exist in the current estimate, leading to redundant scene entities. Although this behavior is uncommon when evaluating on in-distribution validation data, the out-of-distribution nature of real-world data likely causes ambiguity around when the baseline should predict the \texttt{<STOP>} token. The jointly trained model, however, is much more reliable in OOD scenarios and consistently predicts \texttt{<STOP>} at the correct time and avoids this failure mode.

\section{Action Selection Heuristic}

\algrenewcommand\algorithmicrequire{\textbf{Input:}}
\algrenewcommand\algorithmicensure{\textbf{Output:}}

\begin{algorithm*}
\caption{Action Selection Heuristic}
\label{alg:action_selection_heuristic}
\begin{algorithmic}[1] 
\Require Visible layout entity set $E$. Video frame $f$. IoU threshold $\delta$.
\Ensure Action $a$. Selected entity set $S$.

\State Convert $E$ to instance masks to obtain set of instance masks $I_E$.
\State Run Mask2Former~\cite{cheng2021mask2former} on $f$ to obtain set of instance masks $I_M$.
\State Compute all-pairs IoU for $I_M, I_E$, threshold by $\delta$ to obtain matches.

\State $P = \{\}$  \Comment{potential actions}
\For{mask $e$ in $I_E$}
    \State $c = $ class($e$) \Comment{entity class}
    \If{$c =$ wall}
        \If {$e$ is matched with $C \subseteq I_M[c]$ where $|C| \geq 2$}
        \State $P \leftarrow P \cup \{(\texttt{Infill}, \{e\})\}$ \Comment{potentially split wall via \texttt{Infill}}
        \EndIf
    \Else \Comment{door or window}
        \If{$I_M[c] = \emptyset$}
            \State $P \leftarrow P \cup \{(\texttt{Delete}, \{e\})\}$ \Comment{M2F detects no entities of class $c$: \texttt{Delete}}
        \Else 
            \If{no matches for $e$ in $I_M[c]$}
                \State $P \leftarrow P \cup \{(\texttt{Infill}, \{e\})\}$  \Comment{M2F detects an entity of class $c$ but has low overlap with $e$: \texttt{Infill}}
            \EndIf
        \EndIf
    \EndIf
\EndFor
\For{mask $m$ in $I_M$}
    \State $c = $ class($m$) \Comment{entity class}
    \If{$c =$ wall} 
        \If{$m$ is matched with $T \subseteq I_E[c]$ where $|T| \geq 2$}
            \State $P \leftarrow P \cup \{(\texttt{Infill}, T)\}$  \Comment{potentially merge walls in $T$ via \texttt{Infill}}
        \EndIf
    \Else \Comment{door or window}
        \If{$I_E[c] = \emptyset$}
            \State $P \leftarrow P \cup \{(\texttt{Add}, c)\}$  \Comment{\texttt{Add} the class}
        \EndIf
    \EndIf
\EndFor
\State $(a, S) \sim$ Uniform($P$) \Comment{Select a random action}
\State \Return $(a, S)$
\end{algorithmic}
\end{algorithm*}

Here, we detail the action selection heuristic briefly mentioned in Section~\ref{subsubsec:automatic_heuristic_local_corrections}. Algorithm~\ref{alg:action_selection_heuristic} provides the details. In lines 1-2, we compute sets of instance masks $I_E$ by projecting the current scene layout estimate onto the image to create instance masks, and $I_M$ by running Mask2Former~\cite{cheng2021mask2former} (M2F) for walls/doors/windows. We then compute Intersection over Union (IoU) scores in image space between all pairs of instance masks and threshold them by $\delta$. Next, we loop through all masks in $I_E$ to see if they match with masks in $I_M$ (please see lines 5-20 for the details). Similarly, we loop through masks in $I_M$ to see if they match with masks in $I_E$ (please see lines 21-32 for the details). Finally, we randomly sample an action from the list of possible actions in line 33.

We define the following notation for instance mask sets $I$ to select the masks of a certain class: $I[c] = \{x \in I\ |\ \textrm{class}(x) = c\}$.
Note that while Algorithm~\ref{alg:action_selection_heuristic} shows the heuristic for only 1 frame, Project Aria~\cite{aria_white_paper} is a multi-camera system that provides multiple images per capture, thus we run the algorithm over the images (we use the SLAM camera images).

\section{Conceptual Comparison to Other Global Prediction Methods}

Direct quantitative comparison to existing global prediction methods is not applicable in our setting due to the novelty of the local correction task (thus we created a baseline using vanilla SceneScript, i.e. trained only on global predictions). However, we note that our model performs a global prediction as an initial step (which can be quantiatively compared). By incorporating iterative local corrections, our method further refines this initial prediction, making the final refined layout quantitatively better than its initial global prediction.

In this respect, if our model's global prediction outperforms a competitor's global prediction results, we'd expect that our full system would further outperform the competitor quantitatively. We report RoomFormer's~\cite{yue2023connecting} global prediction results on ASE (from~\cite{avetisyan2024scenescript}) for completeness: 85.2 / 79.8 / 72.3 Global Prediction AvgF1, which can be compared to Table~\ref{table:joint_training_ablation_asev1} in the main paper. Note that these numbers differ slightly from~\cite{avetisyan2024scenescript} due to a difference in F1 thresholds as mentioned in Section~\ref{subsec:datasets_metrics_baselines}.
As our model's global prediction is similar to vanilla SceneScript and still outperforms RoomFormer, we expect that refining our global prediction with our local corrections will yield an even larger quantitative gap.

\section{Additional Results for Human-in-the-Loop System}

\subsection{Global Prediction with Accumulated Points}

\begin{figure}[t]
    \centering
    \captionsetup{type=figure}
    \includegraphics[width=\linewidth]{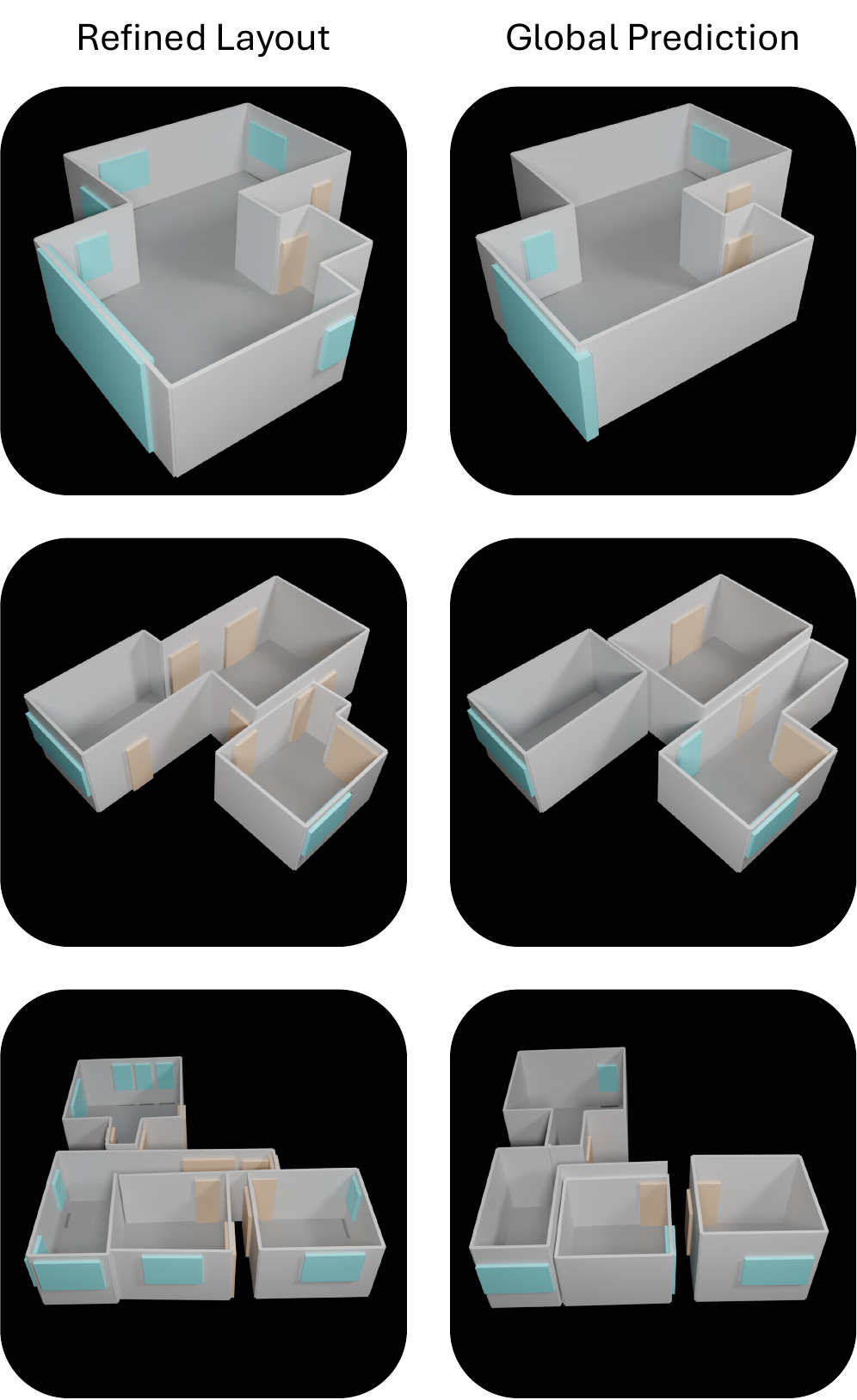}
    \caption{Refined Layout from our Human-in-the-Loop System vs. Global Prediction using all accumulated points. The top row is the scene from Figure~\ref{fig:teaser}, and the bottom two rows are the scenes from Figure~\ref{fig:human-in-the-loop_system} (please reference this figure to see the manually-annotated ground truth for qualitative purposes only). See the text for more details.}
    \label{fig:global_prediction_with_more_points}
\end{figure}

As mentioned in Section~\ref{subsubsec:human-in-the-loop_system}, the human-in-the-loop system accumulates more semi-dense points over time. This allows for users to re-scan poorly scanned areas to obtain more signal for local corrections. In Figure~\ref{fig:global_prediction_with_more_points}, we demonstrate that this additional signal may not be helpful for the global prediction paradigm, likely due to the fact that the ground truth layout is OOD with respect to the training distribution. We show our refined layouts compared to running the model in global prediction mode on the final point cloud with all the accumulated semi-dense points. The global prediction lacks the correct amount and placement of doors and windows, and the wall structure is simplistic in comparison to our refined layout. We hypothesize that these real-world scenes are out of distribution with respect to the training distribution. This again highlights that our human-in-the-loop framework can produce scene layouts that the single-shot global prediction (i.e. vanilla SceneScript) cannot produce.

\subsection{Demo Videos}
At our \href{https://projectaria.com/scenescript}{project webpage}, we provide videos of our human-in-the-loop system running live on a Meta Quest 3 with an Aria device rigidly attached. The three videos correspond to the three real-world scenes used for qualitative results in the main paper (one from Figure~\ref{fig:teaser} and two from Figure~\ref{fig:human-in-the-loop_system}). On the left of the video, we highlight the current layout estimate, action, selected entities, and infilled entities. 

\subsection{Failure Cases}
In our supplemental materials, we provide a video of failure cases of our human-in-the-loop system. We detail them here:

\begin{itemize}
    \item Our model struggles to accurately identify individual windows when multiple small windows are positioned closely together. Groups of windows are often misidentified as a single large window. We posit that this is due to the nature of the synthetic training data. This can also be seen in our qualitative results in Figure~\ref{fig:human-in-the-loop_system}.
    \item Although users typically expect a newly added entity to appear directly in front, our model does not consistently follow this behavior in practice. This is likely due to the heuristics used to select data (e.g. visibility as discussed in Section~\ref{subsubsec:local_correction}). Future work involves curating the data to better align the model's behavior with human expectations.
    \item Lastly, we show an example where the model is not capable of producing the correct geometry for a pair of small windows. We speculate that this style of window is not present anywhere in our synthetic training data. Although the local correction task demonstrates better generalization when used in a human-in-the-loop system compared to the global prediction distribution, it is inherently limited to the learned local distributions. As a result, it cannot predict outcomes for patterns or scenarios that fall outside these distributions.
\end{itemize}

\end{document}